%% file: PaperForReview.tex
\documentclass[10pt,twocolumn,letterpaper]{article}

\usepackage{wacv}              

\usepackage{graphicx}
\usepackage{amsmath}
\usepackage{amssymb}
\usepackage{booktabs}
\usepackage[accsupp]{axessibility} %

\usepackage{url}
\usepackage[dvipsnames]{xcolor}
\usepackage{hyperref}
\hypersetup{
	final,
	breaklinks,
	colorlinks,
	linkcolor={Mahogany},
	citecolor={YellowGreen},
	urlcolor={CadetBlue}
}

\usepackage[capitalize]{cleveref}
\crefname{section}{Sec.}{Secs.}
\Crefname{section}{Section}{Sections}
\Crefname{table}{Table}{Tables}
\crefname{table}{Tab.}{Tabs.}

\def\wacvPaperID{1554} 
\def\confName{WACV}
\def\confYear{2024}

\newcommand{\MethodName}{{DPT}}
\begin{document}

\title{Disentangled Pre-training for Image Matting}

\author{
    Yanda Li\textsuperscript{1}, 
    Zilong Huang\textsuperscript{2}, 
    Gang Yu\textsuperscript{2}, 
    Ling Chen\textsuperscript{1}, 
    Yunchao Wei\textsuperscript{3}, 
    Jianbo Jiao\textsuperscript{4} \\
    \textsuperscript{1}University of Technology Sydney, Australia
    \textsuperscript{2}Tencent \\
    \textsuperscript{3}Beijing Jiaotong University, China
    \textsuperscript{4}University of Birmingham, UK\\
    \{liyanda95,wychao1987\}@gmail.com, 
    \{zilonghuang, skicyyu\}@tencent.com\\ Ling.Chen@uts.edu.au, j.jiao@bham.ac.uk
}
\maketitle

\begin{abstract}

Image matting requires high-quality pixel-level human annotations to support the training of a deep model in recent literature. Whereas such annotation is costly and hard to scale, significantly holding back the development of the research.
In this work, we make the first attempt towards addressing this problem, by proposing a self-supervised pre-training approach that can leverage infinite numbers of data to boost the matting performance.
The pre-training task is designed in a similar manner as image matting, where random trimap and alpha matte are generated to achieve an image disentanglement objective. The pre-trained model is then used as an initialisation of the downstream matting task for fine-tuning.
Extensive experimental evaluations show that the proposed approach outperforms both the state-of-the-art matting methods and other alternative self-supervised initialisation approaches by a large margin. We also show the robustness of the proposed approach over different backbone architectures. Our project page is available at \url{https://crystraldo.github.io/dpt_mat/}.

\end{abstract}

\input{latex/main/intro}

\input{latex/main/related}

\input{latex/main/method}
\input{latex/main/exp}

\input{latex/main/conc}

\paragraph*{Acknowledgements}
Jianbo Jiao is supported by the Royal Society Short Industry Fellowship SIF\textbackslash R1\textbackslash231009.

{\small
\bibliographystyle{ieee_fullname}
\bibliography{egbib}
}

\input{supp}

\end{document}

%% file: latex/main/intro.tex
\section{Introduction}
Image matting has played a predominant role in daily applications in the past few years, \eg online meetings and smartphone applications, referring to extracting the foreground from natural images by predicting an alpha matte. A natural image $\mathcal{I}$ can be represented as a linear fusion of foreground $\mathcal{F}$ and background $\mathcal{B}$ with a weighting parameter $\alpha$ as defined below:
   \begin{equation}\label{eq:mat}
      \mathcal{I} =\alpha \mathcal{F}+(1-\alpha)\mathcal{B}.
      \vspace{-1mm}
  \end{equation}

     \begin{figure}
    \centering
    \includegraphics[width=\columnwidth]{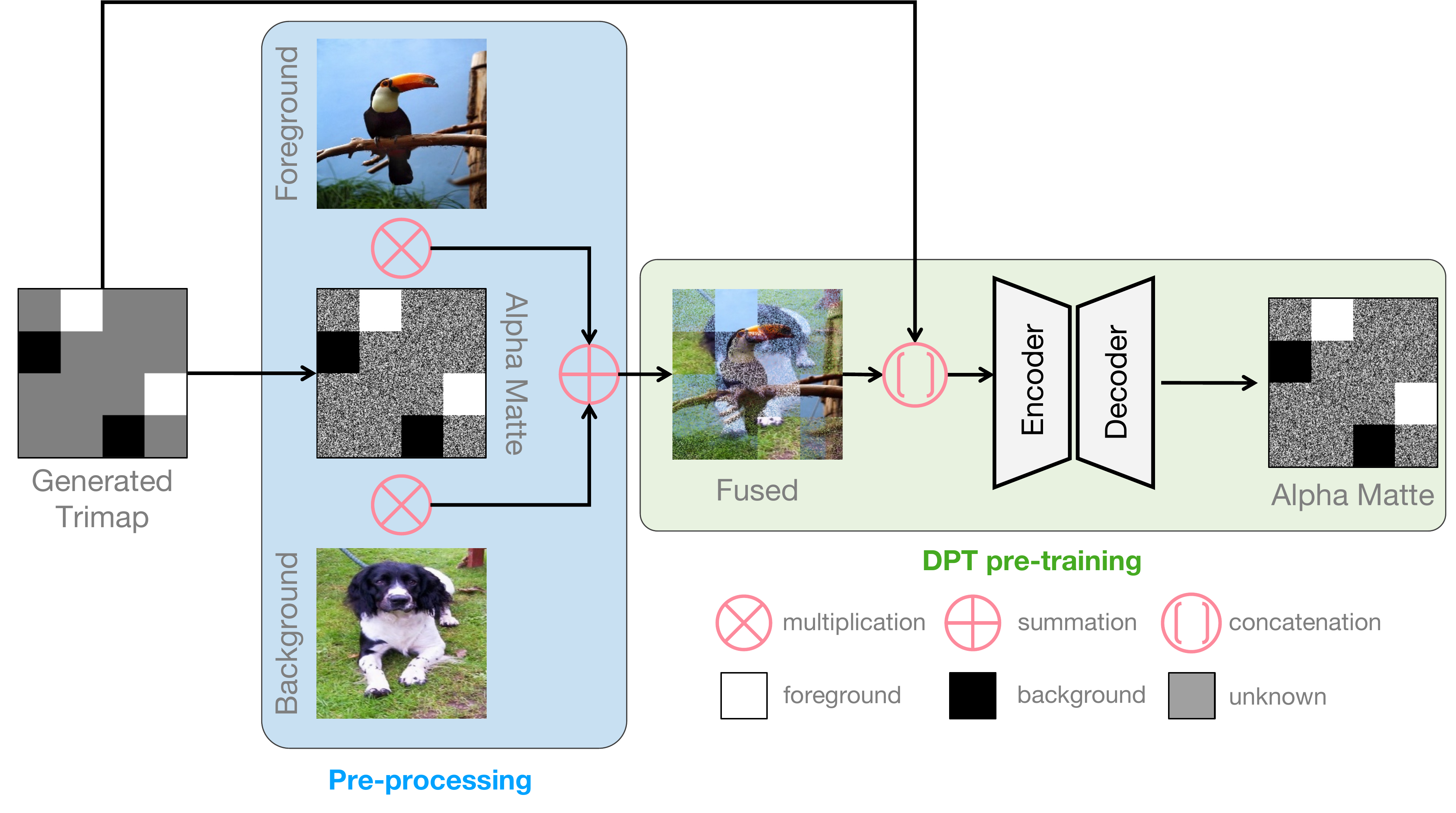}
    \caption{An overview of the proposed \MethodName. During pre-processing, a trimap is generated by assigning regions of foreground (white blocks), background (black blocks) and unknown (the rest grey ones). Based on that, a random alpha matte $\alpha$ is generated. The foreground and background images are composited with the $\alpha$ according to Eq.~\ref{eq:mat}. Together with the trimap as input, the network is trained to estimate an alpha matte. The aforementioned generated $\alpha$ is used as a pseudo label to train the model.}
    \label{fig:tenser}
    \vspace{-5mm}
   \end{figure}
   
Matting, as a low-level computer vision problem, has been studied for decades in the literature and remains a challenging problem.
With the availability of computational resources and network capacity, most existing work addresses this problem based on deep neural networks.
Among these methods, the contributions mainly lie on adjusting the objective function~\cite{lutz2018alphagan, forte2020f, hou2019context}, network structure~\cite{xu2017deep, tang2019learning, qiao2020attention, park2022matteformer} or the data input to the network~\cite{aksoy2018semantic, sengupta2020background, shen2016deep}. 
Due to the data-driven nature of deep learning, the quality of data and its annotation is the key to the model performance, apart from the computational resources, and is becoming the bottleneck.
However, due to the nature of the image matting task and its high-precision requirement for the pixel-level annotation of the alpha matte, data labelling is incredibly costly and hard to scale, restricting the development of the field significantly.
Composition-1k~\cite{xu2017deep} and Distinct-646~\cite{qiao2020attention}, the most commonly used matting datasets, only contain hundreds of foreground images with corresponding annotations, and serve as the main benchmarks for the whole community.
To address this vital and timely challenge, instead of seeking more labour to annotate more data, we step back and are more interested in the question: \textit{is that possible to leverage the freely available huge amounts of unlabelled natural images to boost the performance of image matting?}
To this end, in this work we make the first attempt towards self-supervised pre-training for image matting.

  \begin{figure*}
    \centering
    \includegraphics[width=\textwidth]{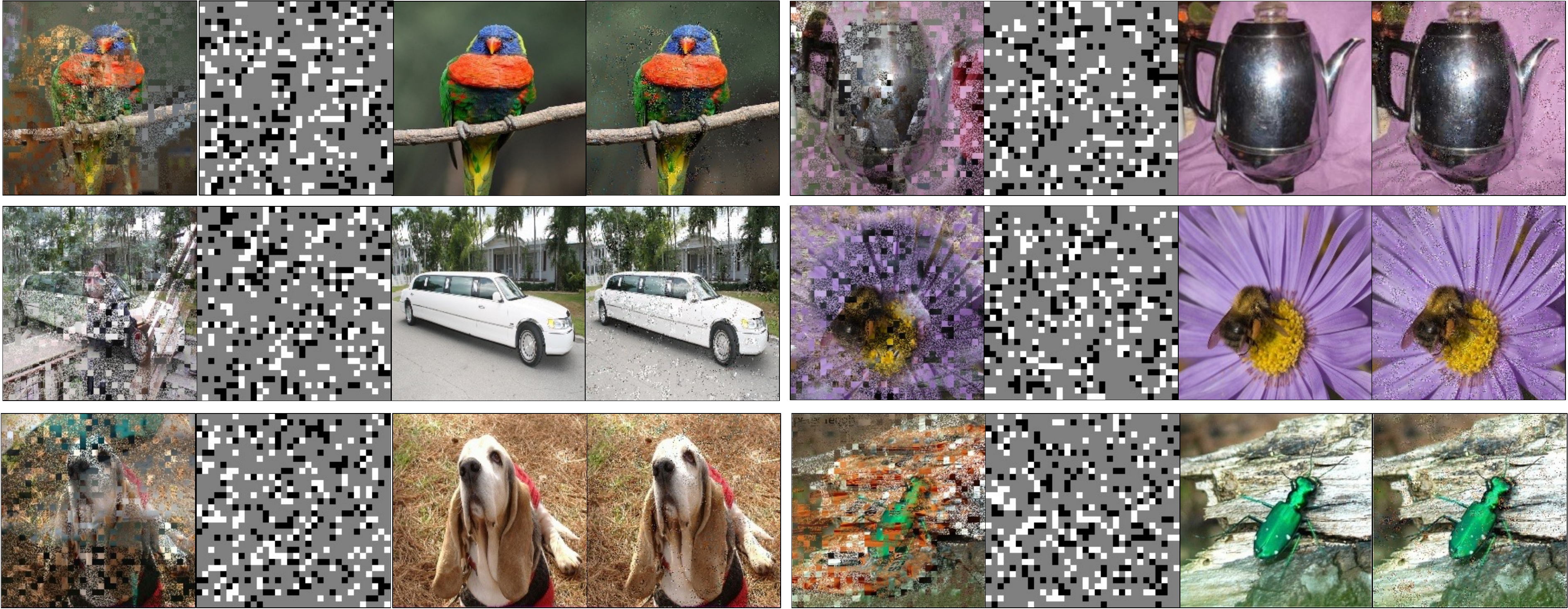}
    \caption{Example results on ImageNet-1k. For each set of images, we show the merged image (left), defined trimap (second), labelled foreground image (third) and generated foreground image (right). The size of patches in trimap is 7$\times$ 7 and the unknown region occupies 75$\%$ of the patches, 12.5$\%$ for foreground and 12.5$\%$ for background. According to Eq.~\ref{eq:mat}, we combine the merged image according to the labelled alpha matte and pseudo label, respectively, to generate labelled foreground and generated foreground. *Following generic trimap-based methods, the model only predicts the unknown region, and both the foreground region and background region are copied directly from the foreground image.}
    \label{imagenet}
    \vspace{-3mm}
 \end{figure*}  
     
Recently, self-supervised learning (SSL) has made prominent progress in deep learning, which benefits from the rapid update of hardware. In natural language processing, self-supervised learning was first proposed to cater for the growth of millions of data. Large-scale language modelling without human-annotated labels has obtained brilliant success by autoregressive language modeling~\cite{radford2018improving} and masked language modeling~\cite{devlin2018bert}. 
The idea of self-supervised learning has also greatly benefited computer vision. Lots of methods~\cite{he2020momentum, he2022masked, xie2021self} are built to enhance the representation learning of models and achieve superior results in downstream tasks. SimCLR~\cite{chen2020simple} and MoCo~\cite{he2020momentum} made the breakthrough in self-supervised learning on high-level image classification representation. Several following-up works~\cite{cao2020parametric, xie2021self} enhanced the ability of representation learning by improving the contrastive loss. Although visual self-supervised representation learning has proved its effectiveness in several high-level tasks, 
low-level tasks have barely been touched and a suitable self-supervised pre-training approach for image matting is under-explored.
To bridge the gap, two challenges need to be addressed:

i) Auxiliary input is required for the matting task, \ie trimap. According to Eq.~\ref{eq:mat}, there are 3 unknown parameters. To estimate $\alpha$ given only $\mathcal{I}$ is a massively ill-posed problem, hence additional guidance is required. 
Existing SSL methods either learn by predicting the missing parts or constructing positive-negative pairs for contrastive learning, without considering additional guidance during the pre-training, making it infeasible to directly apply off-the-shelf SSL methods for the matting task.

ii) Image matting, by definition, is a pixel-level disentanglement problem. It not only requires foreground-background segmentation, but also needs to estimate the transparency of the fused regions near boundaries. In addition, it is a class-agnostic task, different from existing SSL methods focusing on semantic instance discrimination.
   
To address the above challenges, in this paper, we present a new disentangled self-supervised pre-training approach tailored for image matting, termed \emph{\MethodName}. \MethodName~is designed to be a pretext task similar to the matting task, enabling large-scale pre-training for matting-aware representation learning. In our \MethodName, we simulate the matting process including input guidance and supervision information on synthetic datasets which is similar to matting datasets. With the proposed \MethodName~pre-training, the learned representation is forced to be with the potential ability of image disentanglement. Such representations are leveraged to boost the performance of image matting in the following fine-tuning stage. An illustration of the proposed \MethodName~is shown in Figure~\ref{fig:tenser}. First, trimap $\boldsymbol{T}$ is generated by randomly cropping patches and splitting these patches into three categories of \textit{background}, \textit{foreground} and \textit{unknown}. According to the pre-defined region information, alpha matte $\alpha$ is randomly generated as the pseudo label. A randomly chosen foreground image $\mathcal{F}$, and background $\mathcal{B}$ are fused according to $\alpha$. The fused image is then fed into the network together with the trimap $\boldsymbol{T}$ to predict the aforementioned alpha matte. Some qualitative examples of the proposed pre-training task are shown in Figure~\ref{imagenet}.

With the proposed \MethodName~pre-training, we show the potential of leveraging large-scale (infinite) unlabeled data for image matting with demonstrated clear improvement over the state-of-the-art. 
The main contributions of this work are summarised as below:
\begin{itemize}
    \item We propose, to our knowledge, the first self-supervised large-scale pretraining approach for image matting. The pretext task is designed and tailored for the matting task and shown to be effective in learning disentanglement representations.
    \item The proposed DPT pre-training approach is shown to be effective across different network backbones, with consistent performance improvement.
    \item Extensive experimental analysis on several public datasets shows the effectiveness of the proposed method, outperforming existing image matting approaches by a large margin.
\end{itemize}

%% file: latex/main/related.tex
\section{Related works}
   
\paragraph{Natural image matting}

Traditional natural image matting can be summarized into two categories, sampling-based and propagation-based. Sampling-based methods~\cite{chuang2001bayesian,shahrian2013improving,gastal2010shared} collected foreground and background colour samples to generate alpha of unknown region. Propagation-based methods~\cite{levin2007closed, lee2011nonlocal, chen2013knn} used neighbouring pixels and estimated the alpha matte of the unknown region by propagating the alpha from the foreground and background regions.

Until recently, common matting approaches are divided into two parts: trimap-based and trimap-free. Most methods~\cite{lu2019indices, hou2019context, xu2017deep} took trimap as input to provide regional information. DIM~\cite{xu2017deep} introduced the most popular matting dataset Composition-1k and a two-stage encoder-decoder network with trimap as an additional input. Alphagan~\cite{lutz2018alphagan} presented a generative adversarial network for matting to predict alpha. SampleNet ~\cite{tang2019learning} applied sampling methods to the matting task. A novel end-to-end natural image matting method GCA~\cite{li2020natural} was proposed with a guided contextual attention module. External semantic information was incorporated into the model in SIM~\cite{sun2021semantic} to obtain a better alpha matte. As the first transformer-based matting model, Matteformer~\cite{park2022matteformer} introduced a prior token that participates in the self-attention mechanism. TransMatting~\cite{cai2022transmatting} modelled transparent objects with a big receptive field. 
   
For trimap-free methods~\cite{sengupta2020background, li2022referring}, HAttMatting~\cite{qiao2020attention} proposed Distinctions-646
dataset and designed a hierarchical attention structure. Liu et al.~\cite{liu2020boosting} employed coarse annotated data coupled with fine annotated data for semantic human matting without trimaps as an extra input. Background Matting~\cite{sengupta2020background, lin2021real} replaced trimap with an additional background image without the object. MG Matting~\cite{yu2021mask} took coarse mask as its guidance. To control the matting with natural language description, RIM~\cite{li2022referring} proposed a new task named referring image matting to extract object alpha that matched the given language description. While there have been some outstanding trimap-free methods, however, there is a certain gap between trimap-free methods and trimap-based ones in performance. Considering trimap is widely used for image matting, we adopt trimap as the auxiliary input of the disentangled pre-training task.

\paragraph{Self-supervised learning}
Self-supervised learning approaches have been significantly used in deep learning. In natural language processing, self-supervised learning is mainly achieved by auto-regressive language modeling~\cite{radford2018improving,radford2019language} and masked language modeling~\cite{devlin2018bert}. These methods filled the masked portion of the input sequence by predicting the missing part for pre-training. In this way, the pre-trained model could fit hundreds of millions of data generalize well and achieve better performance when fine-tuning downstream tasks. 

Motivated by the success of unsupervised learning in NLP, some self-supervised learning methods~\cite{he2020momentum, xie2021self, he2022masked, o2020unsupervised, wang2022cp2, bao2021beit, chen2020simple, xie2021detco} are introduced for vision tasks.
MoCo~\cite{he2020momentum} presented an unsupervised pre-training and transferred it to various downstream tasks by fine-tuning. MoBY~\cite{xie2021self} transferred adaption to detection and segmentation with the Swin transformer network. Masked autoencoder (MAE)~\cite{he2022masked} randomly masked patches and reconstructed the missing region. A similar idea was also verified in SimMIM~\cite{xie2022simmim}, where the input image was masked with moderately large patch size and gained a better representation. To better adapt for dense prediction task,~\cite{wang2021dense,o2020unsupervised} migrated contrast loss from image-level to pixel-level. CP$^2$~\cite{wang2022cp2} facilitated both image-level and pixel-level representation through a simple copy-paste operation. There are many excellent SSL methods, whereas self-supervised learning designed for image disentanglement is under-explored.

%% file: latex/main/method.tex
\section{Method}
Our \MethodName~ is a simple self-supervised pre-training approach for image matting task. The architecture of our model is an encoder-decoder framework. We pre-trained our model on the ImageNet-1k (IN1K) training set. Unlike the previous SSL methods, we additionally input the guidance to provide the region information required for image matting. Referring to most of the previous matting works, we decided to take widely adopted trimap as the guidance, which delineates the background, foreground, and unknown region. 

In the pre-training process, the trimap and the corresponding alpha matte are randomly generated first, meanwhile, the two natural images are randomly selected as background and foreground, respectively. Then the fused image is composited with the alpha matte, background image, and foreground image via Eq.\ref{eq:mat}.  At last, the fused image and the one-hot trimap are concatenated as the input of the network to predict alpha matte. The randomly generated alpha matte will be used to supervise the predicted alpha matte. After self-supervised pre-training, we fine-tune on downstream matting dataset using the trained model parameters as initialization. 

Next, we will introduce the details of generating the trimap, the corresponding alpha matte, and the objective function.

\vspace{-2mm}
\paragraph{Trimap generation}
Trimap, as the crucial guidance for matting, is defined in advance. In this paper, we adopt a very simple but effective method to generate trimap. A blank trimap $T \in \mathbb{R}^{H\times W}$ is created first, where $H$ and $W$ are the height and width, respectively. The trimap is with the same resolution as the input image. Then $T$ will be divided into grids of the same size, such as $7\times7$ pixels. We randomly select $\theta$ percent of grids as the unknown region, $\beta$ percent of grids as the foreground region, and $\gamma$ percent of grids as the background region, $\theta + \beta + \gamma = 1$. The background is assigned 0, the foreground is assigned 2, and the unknown region is assigned 1. In the training step, the $\boldsymbol{T}$ is converted into a one-hot embedding as the input guidance of the model. The trimap generation is random and is not independent of the content of the input image. Here, we randomly select based on grids rather than pixels to keep the structure of foreground and background. 


   \begin{figure}
    \centering
    \includegraphics[width=\columnwidth]{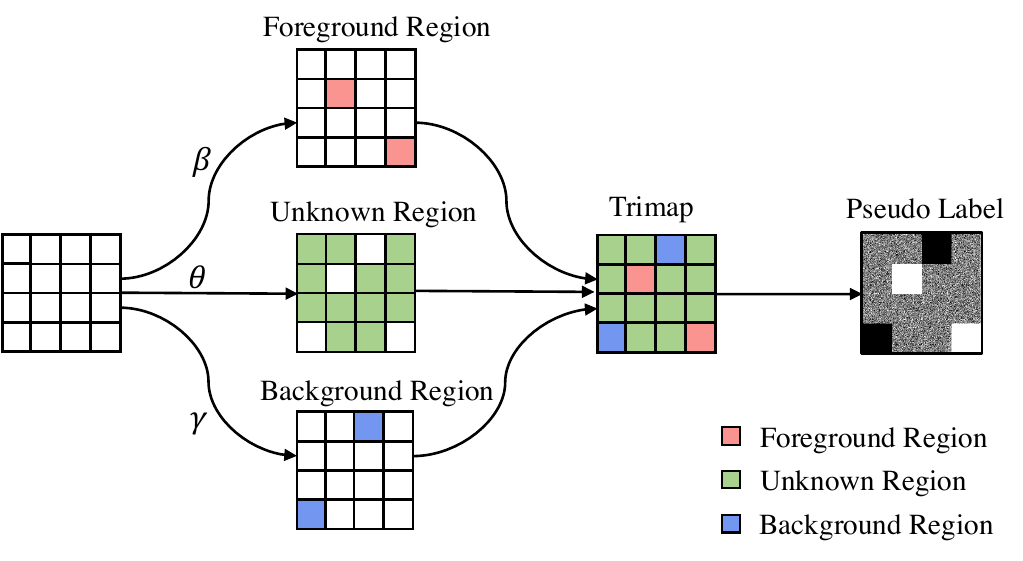}
    \caption{The generation of trimap and pseudo label.~$\theta$ percent of patches in trimap as unknown region, $\beta$ percent for background $\gamma$ percent for foreground, where $\theta + \beta + \gamma = 1$ . Then a random alpha matte $\alpha$ is generated as a pseudo label according to the trimap. }
    \label{fig:trimap}
    \vspace{-3mm}
   \end{figure}

\vspace{-2mm}
\paragraph{Alpha matte generation}
After obtaining trimap, we continue to generate the corresponding alpha matte $\alpha$. Alpha matte is a matrix that has the same size as the trimap. Each value $\alpha_i$ of the alpha matte is in the range of 0 to 1 and will be used to fuse the foreground image and the background image. According to Eq.~\ref{eq:mat}, $\alpha=0$ means the fused image is equal to the background image, $\alpha=1$ means the fused image is equal to the foreground image. 
For each position $i$ in the alpha matte, we set $\alpha_i=1$ if the given trimap $T_i=2$, and $\alpha_i=0$ if the given trimap $T_i=0$. For the position in which $T_i=1$, we randomly assign a number between 0 and 1 for each pixel. It should be emphasized that, in order to be closer to the annotation of the matting dataset, we randomly select an integer value between 0 to 255 for each pixel in the unknown region, then divide by 255 as the final value. The flow chart is shown in Figure~\ref{fig:trimap}. In the experiment section, we discuss different ways to generate trimap and Alpha matte.

\vspace{-2mm}
\paragraph{Loss function}
We adopt three kinds of generic loss functions: L1 regression loss, Composition loss~\cite{xu2017deep} and Laplacian loss~\cite{hou2019context} for both pre-training and fine-tuning stage. The final loss is the sum of the above three losses.
     \begin{equation}
       L_{final} = L_{l1}+L_{comp}+L_{lap},
   \end{equation}
where $L_{l1}$ is the absolute difference between the predicted alpha matte and ground truth. The $L_{comp}$ is the absolute difference between the original image and fused image, where the fused image is generated with the original foreground, original background and predicted alpha matte according to Eq.~\ref{eq:mat}. The $L_{lap}$ calculates differences of two Laplacian pyramid representations between ground truth and predicted alpha matte. All the above losses are calculated only in the unknown region.

\vspace{-2mm}
\paragraph{Discussion} 
Our \MethodName~is similar to natural image matting. We both use additional guidance as an external input to extract the foreground from the fused image. However, there are still some differences.
The normal image matting task has a small number of annotations and increases the amount of training data with synthetic images. The trimap is produced from the ground truth alpha matte by binarizing the foreground objects with a threshold with random dilation. In this case, the unknown region is around the boundaries of the foreground object, which also makes it easier for the model to converge. The training of \MethodName~is based on large-scale unlabeled data. Trimap and alpha matte are randomly generated, which increases the difficulty of training. However, the huge amount of unlabeled data makes up for this shortcoming and greatly improves the final performance.

%% file: latex/main/exp.tex
\section{Experiments}
In this section, we report the detailed setting of our experiments and conduct our experimental evaluations on image matting datasets. In the following, we first compare our method with other state-of-the-art matting methods on Composition-1k~\cite{xu2017deep} and Distinct-646~\cite{qiao2020attention} datasets. Then we compare different alternative pre-training methods with the proposed \MethodName. We also apply our \MethodName~ pre-training on different backbones to validate its robustness. Extensive ablation studies are conducted to validate the contributions of each technical detail.

\subsection{Experimental setting}

\paragraph{Data augmentation}
  We do self-supervised pre-training on the ImageNet-1k (IN1K) training set~\cite{deng2009imagenet}. Following previous pre-training approaches~\cite{he2022masked, liu2021swin}, we use the default image input size of 224$\times$224. For each foreground image, we randomly select an image as the background. The trimap and pseudo label are kept the same size as the image. The patch size within the trimap is set as 7$\times$7. The ratios of an unknown region, foreground, and background are set to 75$\%$, 12.5$\%$ and 12.5$\%$, respectively. We first perform an affine transformation with a random degree, scale, shear, and flip. After that, we randomly change the Hue values of the image. Finally, we composite the foreground and background images with pseudo alpha matte.
  
  In the fine-tuning stage, we perform our experiments on image matting datasets. The data augmentation is set following MG Matting~\cite{yu2021mask}. Two foreground images are randomly fused with alpha matte, followed by random affine transformations and colour jitterings. Patches with size 512$\times$ 512 are cropped around the unknown area in the central area of the foreground image. The augmented foreground and background are then fused according to the ground truth alpha matte.

\vspace{-2mm}
\paragraph{Pre-training} We employ the AdamW optimizer with $\beta_1$=0.9 and $\beta_2$=0.95 for our objective functions. The learning rate adopts a cosine annealing strategy with the base learning rate of $3\times10^{-4}$ and a weight decay of 0.05. By default, we perform self-supervised pre-training with batch size 512 on a single machine equipped with eight NVIDIA V100 GPUs. The model is trained for 100 epochs with a 10-epoch linear warm-up stage.
\vspace{-2mm}
\paragraph{Fine-tuning} In the fine-tuning stage, we follow the same setting as in MatteFormer~\cite{park2022matteformer}. We initialize the network with the Tiny model of
Swin Transformer~\cite{liu2021swin} pretrained on IN1K. The input size of the network is 512$\times$ 512, batch size of 40 for four V100 GPUs on one machine. We initialize the base learning rate with $10^{-3}$. Adam optimizer is adopted with $\beta_1$=0.5 and $\beta_2$=0.999 for training 200k iterations. In the first 5k iterations, we warm up the learning rate by linear increasing to help the model convergence. 

\vspace{-3mm}
\paragraph{Evaluation metrics}
Following the common practice in previous matting methods~\cite{qiao2020attention, yu2021mask,park2022matteformer}, we adopt the sum of absolute differences (SAD), mean squared error (MSE), gradient (Grad), and connectivity errors (Conn) as our evaluation metrics. The fused image and 3-channel trimap are concatenated in the channel dimension as inputs for the networks. The input image is padded to a size of multiple of 32 to facilitate the downsampling of the model and afterwards restored to its original size for evaluation. Note that lower values of the four evaluation metrics indicate higher performance (\ie more accurate alpha matte estimation).

\subsection{Quantitative comparison to state-of-the-arts}
\paragraph{Composition-1k}
Here we test our \MethodName~on Composition-1k~\cite{xu2017deep} test set and compare it with state-of-the-art approaches. Composition-1k is a synthetic matting dataset that contains 431 foreground objects and corresponding labelled alpha matte for training. The test set contains 50 foreground objects that are composited with 20 background images chosen from Pascal-VOC~\cite{everingham2010pascal}, a total of 1,000 samples. The proposed \MethodName~surpasses previous methods with a large margin as shown in Table~\ref{tab:exp_1k}. 
Surprisingly, our \MethodName~outperforms MatteFormer by a large margin, by simply replacing its pre-training weights from IN1K supervised one to our proposed self-supervised one, suggesting the effectiveness of the proposed approach.

\begin{table}[t]
\caption{Quantitative fine-tuning results on Composition-1K. *~indicates additional semantic information as input.}
\centering
\resizebox{\columnwidth}{!}{
\begin{tabular}{l|cccc}
\toprule
Method &  
  {SAD$\downarrow$} &
  {MSE ($10^{-3}$)$\downarrow$} &
  {Grad$\downarrow$} &
  {Conn$\downarrow$} \\ 
  \hline
  KNN-Matting\cite{chen2013knn} & 175.4 & 103.0 & 124.1 & 176.4 \\
  DIM\cite{xu2017deep} & 50.4 & 14.0 & 31.0 & 50.8 \\
  AlphaGAN\cite{lutz2018alphagan} & 52.4 & 30.0 & 38.0 & - \\
  IndexNet\cite{lu2019indices} & 45.8 & 13.0 & 25.9 & 43.7 \\
  SampleNet\cite{tang2019learning}  & 40.4 & 9.9 & - & - \\
  Context-Aware\cite{hou2019context} & 35.8 & 8.2 & 17.3 & 33.2 \\
  GCA\cite{qiao2020attention} & 35.3 & 9.1 & 16.9 & 32.5 \\
  HDMatt\cite{yu2021high} & 33.5 & 7.3 & 14.5 & 29.9 \\
  TIMI-Net~\cite{liu2021tripartite} &29.1 & 6.0 & 11.5 & 25.4 \\
  MG Matting\cite{yu2021mask} & 32.1 & 7.0 & 14.0 & 27.9 \\
  SIM*\cite{sun2021semantic} & 28.0 & 5.8 & 10.8 & 24.8 \\
  RMat~\cite{dai2022boosting} & 25.0 & - & 9.0 & - \\
  MatteFormer~\cite{park2022matteformer} & 23.8 & 4.0 & 8.7 & 18.9 \\
  TransMatting~\cite{cai2022transmatting} & 26.8 & 5.2 & 10.6 & 22.1 \\
  \hline
  \MethodName~(Ours) & \textbf{21.0} & \textbf{3.1} & \textbf{7.0} & \textbf{15.9}
 \\
 \bottomrule
\end{tabular}
\label{tab:exp_1k}
}
\vspace{-3mm}
\end{table}

   \begin{figure*}
    \centering
    \includegraphics[width=\textwidth]{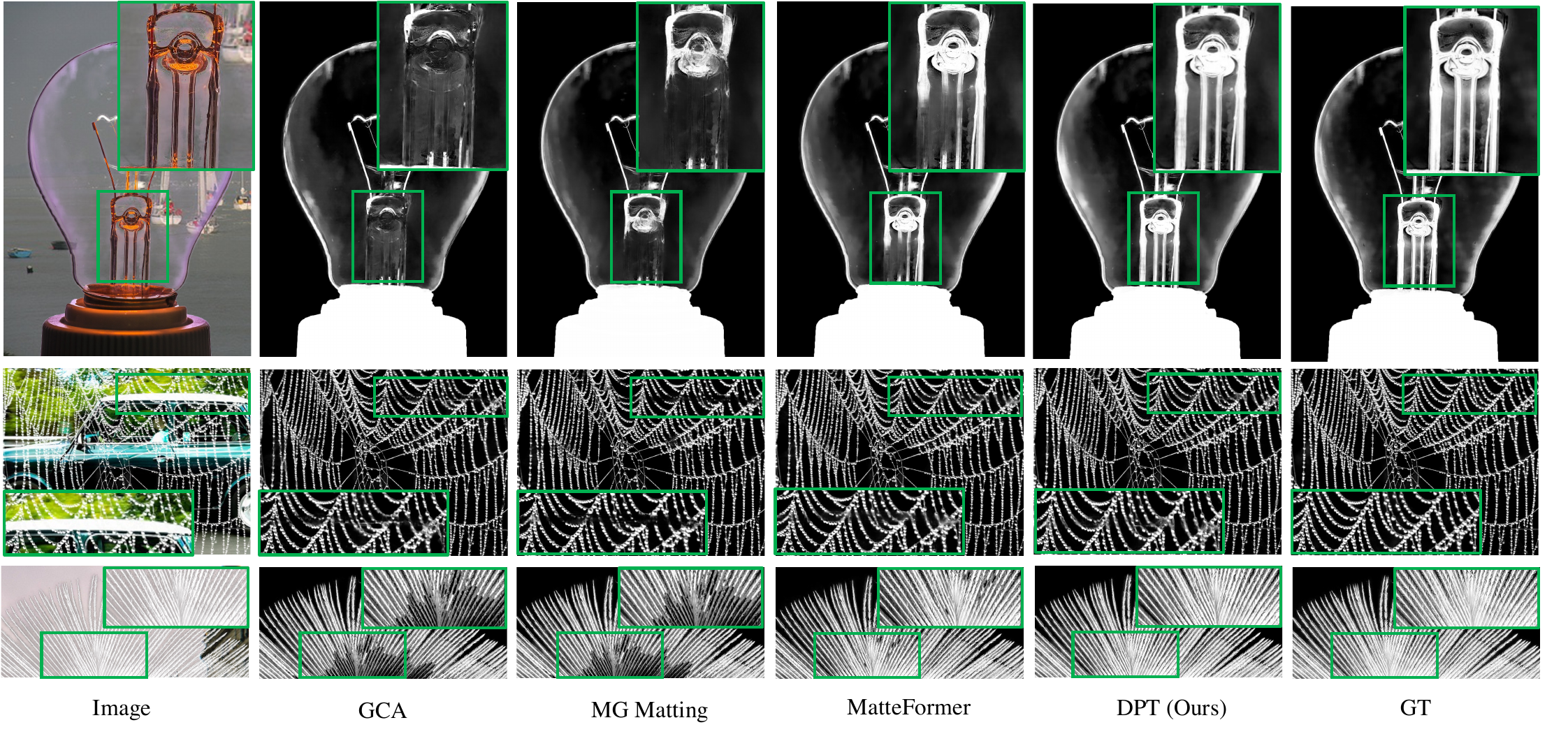}
    \caption{The qualitative results of ours and other state-of-the-art methods on Composition-1k test set.}
    \label{fig:qual}
   \end{figure*}

\paragraph{Distinct-646}
Distinct-646~\cite{qiao2020attention} is a matting benchmark dataset containing 59.6k training images and 1k test images, in total 646 distinct foreground alpha mattes. Unlike the previous matting dataset, it does not provide trimap annotations or other guidance, hence it is difficult to make a fair comparison with other methods. Therefore, we generate trimap by randomly dilating alpha mattes from the ground truth alpha matte as done in MG Matting~\cite{yu2021mask}.

After obtaining the trimap guidance, we compare with the state-of-the-art methods and report the performance in Table~\ref{tab:exp_646}, in which the methods marked with $^*$ are trained on Composition-1k, while others are trained on Distinct-646. For a fair comparison, here we compare with MG Matting and MatteFormer by using the exact same testing setting and test on the Distinct-646 test set. It can be seen from the results that the proposed \MethodName~outperforms other methods with a clear margin across all four evaluation metrics.

\begin{table}[t]
\caption{Quantitative results on Distinct-646. Results with $^*$ are only trained on Composition-1k, others are trained on Distinct-646 as reported in HAtt Matting\cite{qiao2020attention}.}.
\centering
\resizebox{\columnwidth}{!}{
\begin{tabular}{l|cccc}
\toprule
Method &  
  {SAD$\downarrow$} &
  {MSE ($10^{-3}$)$\downarrow$} &
  {Grad$\downarrow$} &
  {Conn$\downarrow$} \\ 
  \midrule
  Closed-Form~\cite{levin2007closed}  & 105.7 & 23.0 & 91.8 & 114.6 \\
  Learning Based~\cite{zheng2009learning}  & 105.0 & 21.0 & 94.2 & 110.4 \\
  KNN Matting~\cite{chen2013knn} & 116.7 & 25.0 & 103.2 & 121.5 \\
  DIM~\cite{xu2017deep} & 47.6 & 9.0 & 43.3 & 55.9 \\
  HAttMatting~\cite{qiao2020attention} & 49.0 & 9.0 & 41.6 & 50.0 \\
  \midrule
  DIM$^*$~\cite{xu2017deep} & 48.7 & 11.2 & 42.6 & 50.0 \\
  Index$^*$~\cite{lu2019indices} & 47.0 & 9.4 & 40.6 & 46.8 \\
  Context-Aware$^*$~\cite{hou2019context} & 36.3 & 7.1 & 29.5 & 35.4 \\
  GCA$^*$~\cite{qiao2020attention} & 39.6 & 8.2 & 32.2 & 38.8 \\
  MG Matting$^*$ ~\cite{yu2021mask} & 36.9 & 6.3 & 35.0 & 22.3 \\
  MatteFormer$^*$ ~\cite{park2022matteformer} & 31.0 & 4.9 & 22.6 & 19.6 \\
  \midrule
  \MethodName (Ours)~$^*$ & \textbf{28.5} & \textbf{3.8} & \textbf{18.0} & \textbf{19.0} \\
  \bottomrule
\end{tabular}
}
\label{tab:exp_646}
\end{table}

\subsection{Qualitative performance}
The qualitative results of ours and other state-of-the-art methods on the Composition-1k test set are shown in Figure~\ref{fig:qual}, in which we compare with GCA~\cite{qiao2020attention}, MG Matting~\cite{yu2021mask} and MatteFormer~\cite{park2022matteformer}. In most samples, various methods can achieve good results, but in some challenging cases (\eg cobweb, light bulb), our method performs much better, especially in detailed regions.


\subsection{Analysis on alternative pre-training methods}
To validate the effectiveness of the proposed pre-training approach, we consider several representative existing pre-trained models developed for Swin Transformer (Tiny model). The classification supervised pre-training and Transformer-SSL are pre-trained for 300 epochs, SimMIM and \MethodName~are trained for 100 epochs. All the methods are pre-trained on IN1K~\cite{deng2009imagenet} only, and the pre-trained models are used as initialisation of MatteFormer for fine-tuning on Composition-1k. The results are shown in Table~\ref{tab:pre-train}:

\begin{itemize}
\setlength{\itemindent}{0em}
\item \emph{Random}: without any pre-trained weights for initialisation, we randomly initialize the model parameters, and directly train from scratch.
\item \emph{Supervised}: the network initializes with the most commonly used IN1K~\cite{deng2009imagenet} fully-supervised pre-trained model.
\item \textit{SimMIM}~\cite{xie2022simmim}: a simple framework for SSL by masked image modeling.
\item \textit{MoBY}~\cite{xie2021self}: an SSL method combined with contrastive loss, serving as a representative contrastive learning-based work.
\end{itemize}
\begin{table}[t]
\caption{With different pre-training methods, quantitative results after fine-tuning on Composition-1k test set.}
\centering
\resizebox{0.9\columnwidth}{!}{
\begin{tabular}{l|cc}
\toprule
Pre-train Method &  
  {SAD}$\downarrow$ &
  {MSE ($10^{-3}$)}$\downarrow$ \\
  \midrule
  Random & 49.2 & 14.7 \\
  Supervised~\cite{deng2009imagenet} & 23.8 & 4.0 \\
  SimMIM~\cite{xie2022simmim}  & 24.6 & 4.2 \\
  Transformer-SSL (MoBY)~\cite{xie2021self}  & 24.0 & 3.9 \\
  \MethodName~ (Ours) & \textbf{21.0} & \textbf{3.1} \\
  
 \bottomrule
\end{tabular}
\label{tab:pre-train}
}
\end{table}

It can be seen from the experimental results that the performance of the model without the pre-trained weight is much worse than those with the pre-trained weight, suggesting that due to the limited number of labelled matting images, prior knowledge needs to be obtained and is indeed helpful. By using pre-trained weights (second row) it achieves a positive performance gain on the matting task. The remaining two self-supervised methods can achieve comparable performance with fully supervised classification. Surprisingly, our proposed \MethodName~achieves much better performance, even outperforming the supervised counterpart. This verifies that the proposed \MethodName~learns better representations for the matting task.

\begin{table}
\caption{With different backbone pre-training, fine-tuning quantitative results on Composition-1k test set.}
\centering
\resizebox{\columnwidth}{!}{
\begin{tabular}{l|ccccc}
\toprule
Method & Backbone & Init. & {SAD}$\downarrow$ & {MSE ($10^{-3}$)}$\downarrow$ \\
  \midrule
  GCA~\cite{qiao2020attention} & Resnet-34 & Supervised & 35.3 & 9.1 \\
  GCA~\cite{qiao2020attention} & Resnet-34 & DPT & \textbf{33.0} & \textbf{8.3} \\
  \midrule
  MG Matting~\cite{yu2021mask} & Resnet-34 & Supervised & 29.3 & 6.3 \\
  MG Matting~\cite{yu2021mask} & Resnet-34 & DPT &\textbf{27.1} & \textbf{5.5} \\
  \midrule
  MatteFormer~\cite{park2022matteformer} & Swin-T & Supervised & 23.8 & 4.0 \\
  MatteFormer~\cite{park2022matteformer} & Swin-T & DPT & \textbf{21.0} & \textbf{3.1} \\
  
 \bottomrule
\end{tabular}
\label{tab:backbone}
}
\end{table}
  
\subsection{Analysis on different backbones}
Preceding deep learning-based matting approaches can be roughly divided into two categories in terms of backbone architecture: CNN-based and Transformer-based. Most of the matting works use CNN-based backbones, mainly based on the ResNet~\cite{he2016deep} architecture. With the rise of Swin Transformer~\cite{liu2021swin}, there are also some works based on Transformers. To analyze the scalability of \MethodName, we conduct experiments by pre-training on both backbones and fine-tuning on the same matting dataset. 


For CNN-based approaches, we choose GCA~\cite{qiao2020attention} and MG Matting~\cite{yu2021mask}. The backbone used by GCA and MG Matting is ResNet-34~\cite{he2016deep}. However, the input guidance of MG Matting is a 1-channel mask, different from trimap-based methods. To better compare with it, we replace its input mask with a 3-channel trimap and retrain the model for testing. The corresponding network layer is also modified accordingly. We use the IN1K classification supervised pre-trained model as the baseline. And compare it with GCA and MG Matting by re-training it using our \MethodName.

On the other hand, we select a state-of-the-art Transformer-based approach, MatteFormer~\cite{park2022matteformer}. As above, we also use the weights of classification supervision as the baseline, and re-train MatteFormer with our \MethodName. We load the two types of pre-training parameters separately and perform fine-tuning on the MatteFormer. The performance comparison is shown in Table~\ref{tab:backbone}. 


  

Note that in the fine-tuning stage, only the pre-trained weights have been changed, but we can see a substantial improvement has been achieved on both backbones when using our \MethodName. This validates the robustness and transferability of the proposed method.

\subsection{Ablation study}
\begin{figure}
    \centering
    \includegraphics[width=\columnwidth]{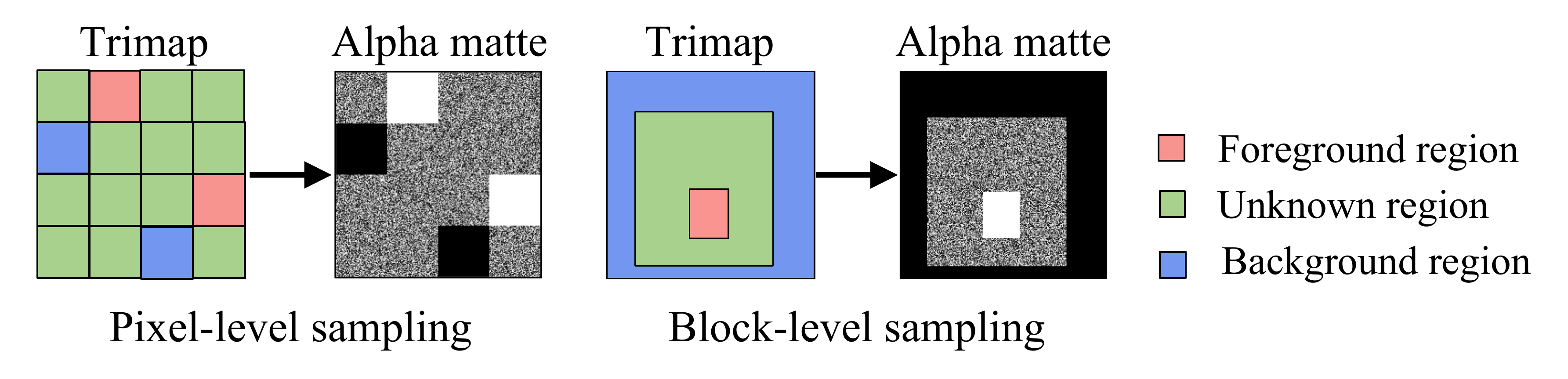}
    \caption{Pseudo alpha matte generation strategy.}
    \label{fig:strategy}
    \vspace{-3mm}
  \end{figure}

\paragraph{Pseudo alpha matte generation strategy}
We compare two pseudo alpha matte generation strategies, as illustrated in Figure~\ref{fig:strategy}. 
\emph{1) Pixel-level sampling strategy}. We first divide a 224$\times$224 image into 7$\times$7 patches. For each patch, it is randomly assigned as either foreground, background or unknown regions, and the trimap is generated consequently. Then pseudo alpha matte is randomly generated in each pixel according to the pre-defined trimap.

The results of the two strategies are shown in Table~\ref{tab:pseudo}. After experiments, both strategies can achieve new state-of-the-art results. We finally chose the pixel-level sampling strategy with slightly higher performance.

\begin{table}
\centering
\caption{Results of different pseudo alpha matte strategies.}
\label{tab:pseudo} 
\begin{tabular}{l|cc}
\toprule
Strategy &  
  {SAD}$\downarrow$ &
  {MSE ($10^{-3}$)}$\downarrow$ \\
  \hline
  Pixel-level sampling & 21.0 & 3.1 \\
  Block-level sampling  & 21.0 & 3.2\\ 
 \bottomrule
\end{tabular}
\end{table}

\vspace{-3mm}
\paragraph{Stages of loading pre-trained weights}
In order to verify the impact of loading pre-trained weights onto different stages on performance, we conduct experiments based on MatteFormer~\cite{park2022matteformer}. We compare the supervised method which only loads the pre-trained weights in the encoder, with \MethodName~loading the pre-trained weights in either the encoder or the whole model. As shown in Table~\ref{tab:stage}, our DPT outperforms the baseline by a large margin, which can be further improved when including the pre-trained decoder.

\begin{table}
\centering
\caption{With different stages of loading \MethodName pre-trained weights, fine-tuning quantitative results on Composition-1k test set.}
\resizebox{\columnwidth}{!}{
\begin{tabular}{l|cccc}
\toprule
Method &  
  Stage & 
  Init. &
  {SAD}$\downarrow$ &
  {MSE ($10^{-3}$)}$\downarrow$ \\
  \midrule
  MatteFormer & Encoder & Supervised & 23.8 & 4.0 \\
  MatteFormer & Encoder  & DPT & 22.1 & 3.4 \\
  MatteFormer & Encoder+Decoder & DPT & {21.0} & {3.1} \\
\bottomrule
\end{tabular}
}
\label{tab:stage}
\end{table}

\paragraph{Impact of trimap for pre-training}
Meanwhile, we perform an ablation study to validate the impact of trimap for the pre-training stage. Unlike \MethodName, the input only contains synthetic images with three channels while the data processing and subsequent fine-tuning stage remain in the same setting. The result is shown as Table~\ref{tab:with}. Although descent results can be achieved without trimap, adding trimap as an additional input further improves performance significantly.

\begin{table}
\centering
\caption{The fine-tuning quantitative results on Composition-1k test set, when with and without trimap during pre-training.}
 \resizebox{0.8\columnwidth}{!}{
\begin{tabular}{l|ccc}
\toprule
Method &  
  Trimap & 
  {SAD}$\downarrow$ &
  {MSE ($10^{-3}$)}$\downarrow$ \\
  \midrule
  MatteFormer & Without & 23.4 & 3.7 \\
  MatteFormer & With  & 21.0 & 3.1 \\
\bottomrule
\end{tabular}
 }
\label{tab:with}
\vspace{-3mm}
\end{table}

\begin{figure}
    \centering
    \includegraphics[width=\columnwidth]{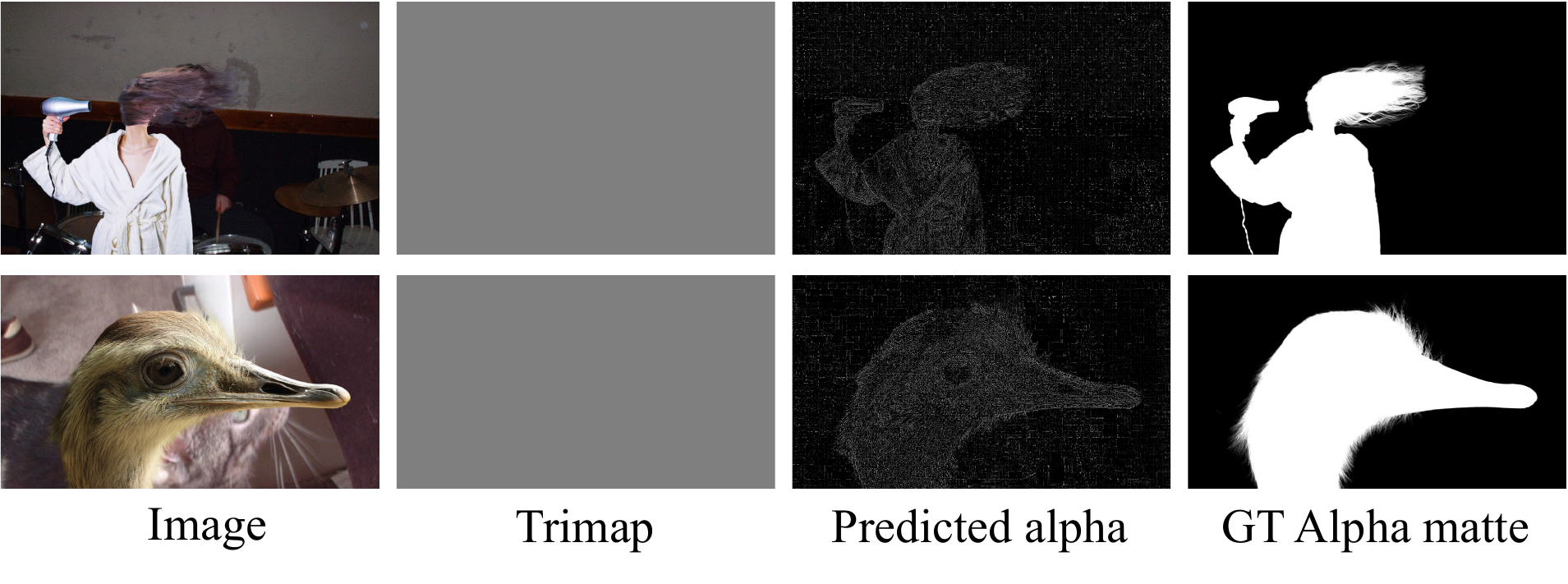}
  \vspace{-6mm}
    \caption{Qualitative performance of directly predicting alpha matte using our pre-trained model with an \textit{All-unknown} trimap.}
    \label{fig:pre}
  \vspace{-2mm}
  \end{figure}

\paragraph{Contour information validation for pre-training}
In order to validate the effectiveness of our pre-training approach on learning contour information, we directly applied the pre-trained model on the comp-1k test set, without fine-tuning. We set \textit{all} input trimaps as \textit{unknown}, \ie without any form of guidance. Then we fed the merged image along with such \textit{All-unknown} trimap into the pre-trained model to predict the alpha. As shown in Figure~\ref{fig:pre}, although without any guidance and fine-tuning, our pre-trained model is able to extract object contour information (even hair and fur), suggesting its capacity to learn and understand high-level object information. This study further validates the effectiveness of our approach for the matting tasks.

\begin{figure}
    \centering
    \includegraphics[width=0.8\columnwidth]{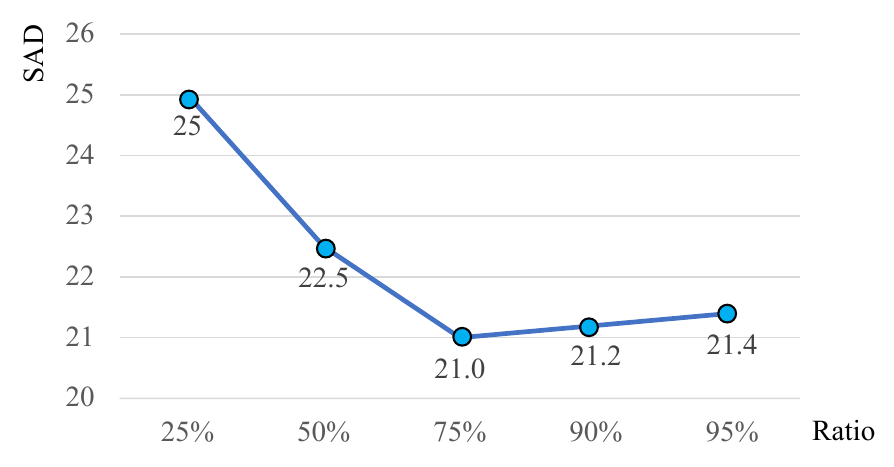}
    \caption{SAD for different ratios of the unknown region in pre-defined trimap. }
    \label{fig:ratio}
   \end{figure}

\paragraph{The ratio of the unknown region in trimap}
The unknown region is necessary for the loss computation. As the size of the unknown region changes, the final results also change. The foreground and background regions also provide information for the network. Thus a suitable ratio of the unknown region will affect the final performance. Here we analyze different ratios of this unknown region and report the results in Figure~\ref{fig:ratio}. It can be seen that when we choose a lower ratio such as 25$\%$ or 50$\%$, the results will be slightly worse. If we choose a higher threshold, such as 75$\%$ or higher, then the final performance will become stable and get better results. As a result, in this paper, we choose the ratio of 75$\%$.

\begin{figure}
    \centering
    \includegraphics[width=\columnwidth]{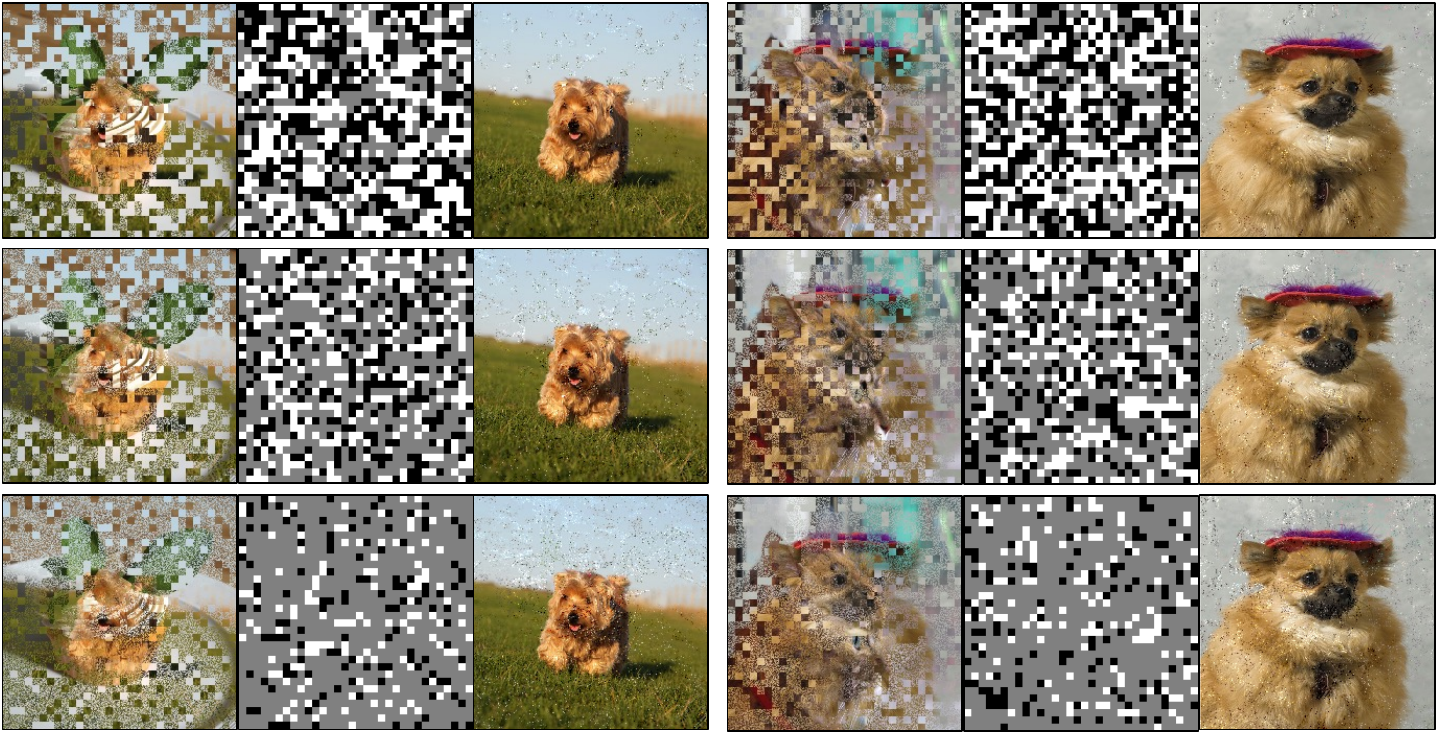}
    \caption{Qualitative results of different unknown ratios of trimap. \textbf{Top to bottom}: unknown region accounts for 25$\%$, 50$\%$ and 75$\%$, respectively. For each image group, from \textbf{left to right}: the merged image, trimap, and the foreground image generated by Eq.~\ref{eq:mat}.}
    \label{fig:block}
   \end{figure}

Visualization of different ratios of the unknown regions in trimap is shown in Figure~\ref{fig:block}. As the unknown area increases, the model is supposed to predict more pixels, and the noise points on the prediction result will also increase, which is revealed in the result. 

%% file: latex/main/conc.tex
\section{Conclusion}
In this work, we looked into the fundamental problem of image matting (\ie data) and proposed a disentangled self-supervised pre-training method named \MethodName. It is designed for the image matting task to enable the utilization of large-scale unlabeled data. More specifically, we generated trimap as an auxiliary input and pseudo alpha matte for supervision, then trained the model towards disentanglement of the fused image. After this pre-training, the derived model was used for the initialization of the downstream image matting task, to boost its performance. Extensive experimental analysis showed the effectiveness and robustness of our \MethodName. We hope this work could attract attention from the community to think about the data leverage side for the matting task and potentially inspire follow-up research.

%% file: supp.tex
\section{Limitations}
The training data we used is class-agnostic, and as a result no semantic information was considered explicitly (though maybe implicitly), which may restrict the potential of the proposed framework to be applied to other semantic-related tasks.
\section{Comparison with the Dense-CL Method}
To verify our effectiveness, we compare our \MethodName~with DenseCL~\cite{wang2021dense} based on GCA~\cite{qiao2020attention}. Since the backbone of GCA is different from that of DenseCL, for a fair comparison, we replace the backbone with the ResNet50, and perform fine-turning after loading pre-trained weights from DenseCL. As shown in Table~\ref{tab:densecl}, our \MethodName~achieves better performance with fewer parameters.

\begin{table}[h]
\caption{Compare with DenseCL based on Comp-1k.}
\centering
\resizebox{\columnwidth}{!}{
\begin{tabular}{l|ccccc}
\toprule
Method &  
Init. &
Backbone &
  {SAD}$\downarrow$ &
  {MSE ($10^{-3}$)}$\downarrow$ \\
  \midrule
  GCA & DenseCL & ResNet50 & 45.9 & 13.0 \\
  GCA & Supervised & ResNet34 & 35.3 & 9.1 \\
  GCA & DPT (Ours) & ResNet34 & \textbf{33.0} & \textbf{8.3} \\
\bottomrule
\end{tabular}
}
\vspace{-4mm}
\label{tab:densecl}
\end{table}

\section{Ablation on losses}
In our \MethodName, we adopt three kinds of losses, L1 regression loss, Composition loss and Laplacian loss. The impact of these loss functions (\textit{$L_{l1}$}, \textit{$L_{lap}$} and \textit{$L_{comp}$}) on the final performance is shown in Table~\ref{tab:loss}.

\begin{table}[h]
\caption{Ablation on losses (50 epochs based on MatteFormer).}
\centering
\resizebox{\columnwidth}{!}{
\begin{tabular}{l|lcc}
\toprule
Method & Loss & {SAD}$\downarrow$ & {MSE ($10^{-3}$)}$\downarrow$ \\
  \midrule
  MatteFormer & $L_{l1}$ & 21.9 & 3.6 \\
  MatteFormer & $L_{l1}+L_{lap}$ & 22.5 & 3.3 \\
  MatteFormer & $L_{l1}+L_{comp}$ & 22.2 & 3.5 \\
  MatteFormer & $L_{l1}+L_{lap}+L_{comp}$ & \textbf{21.0} & \textbf{3.2} \\
\bottomrule
\end{tabular}
}
\vspace{-4mm}
\label{tab:loss}
\end{table}
\vspace{-3mm}

\section{More fine-tuning qualitative results on Composition-1k}
\begin{figure*}
  \centering
  \includegraphics[width=\textwidth]{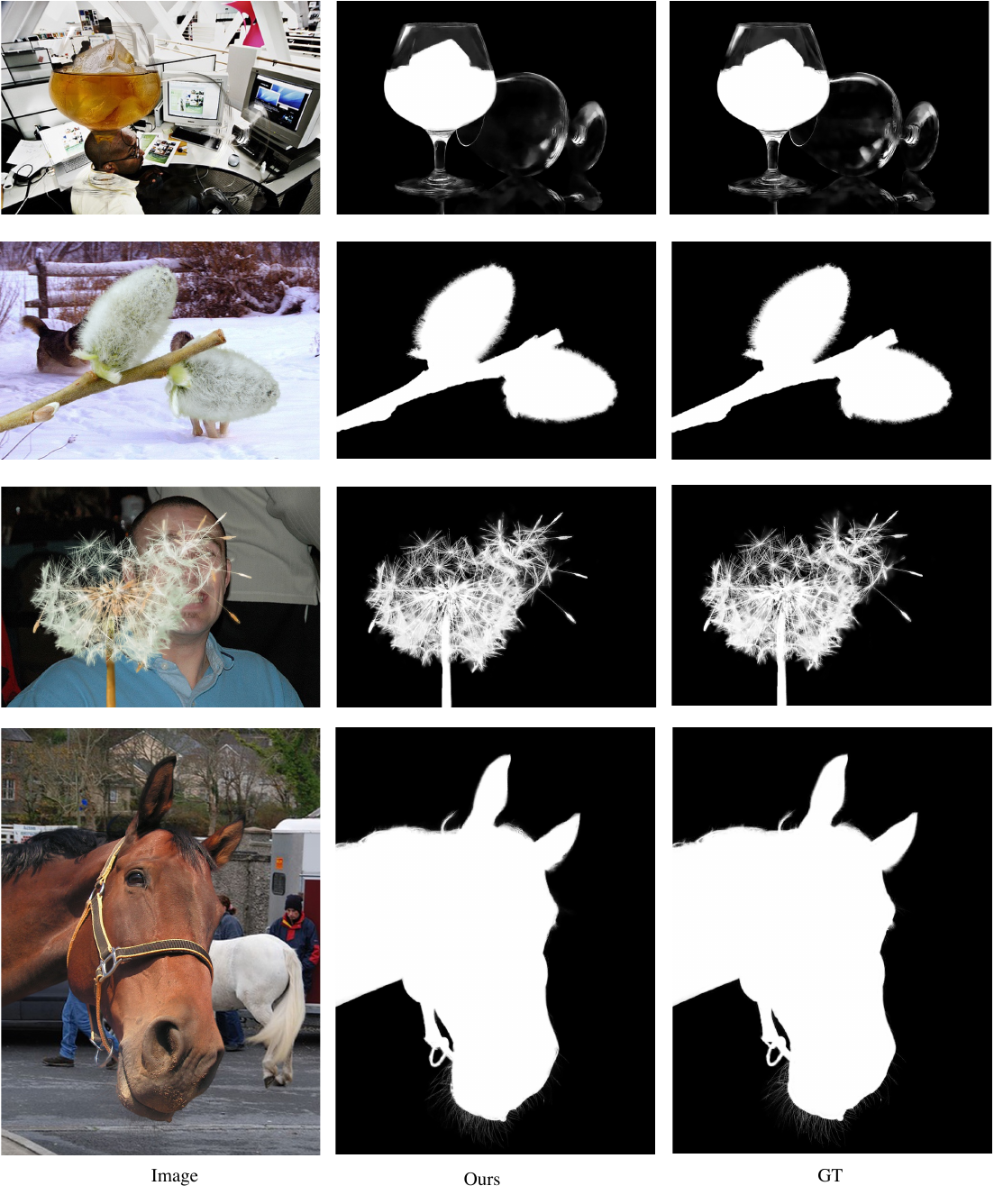}
  \caption{Qualitative results of our method on the Composition-1k test set.}
    \label{comp_supp1.examp}
 \end{figure*}

\begin{figure*}
  \centering
  \includegraphics[width=\textwidth]{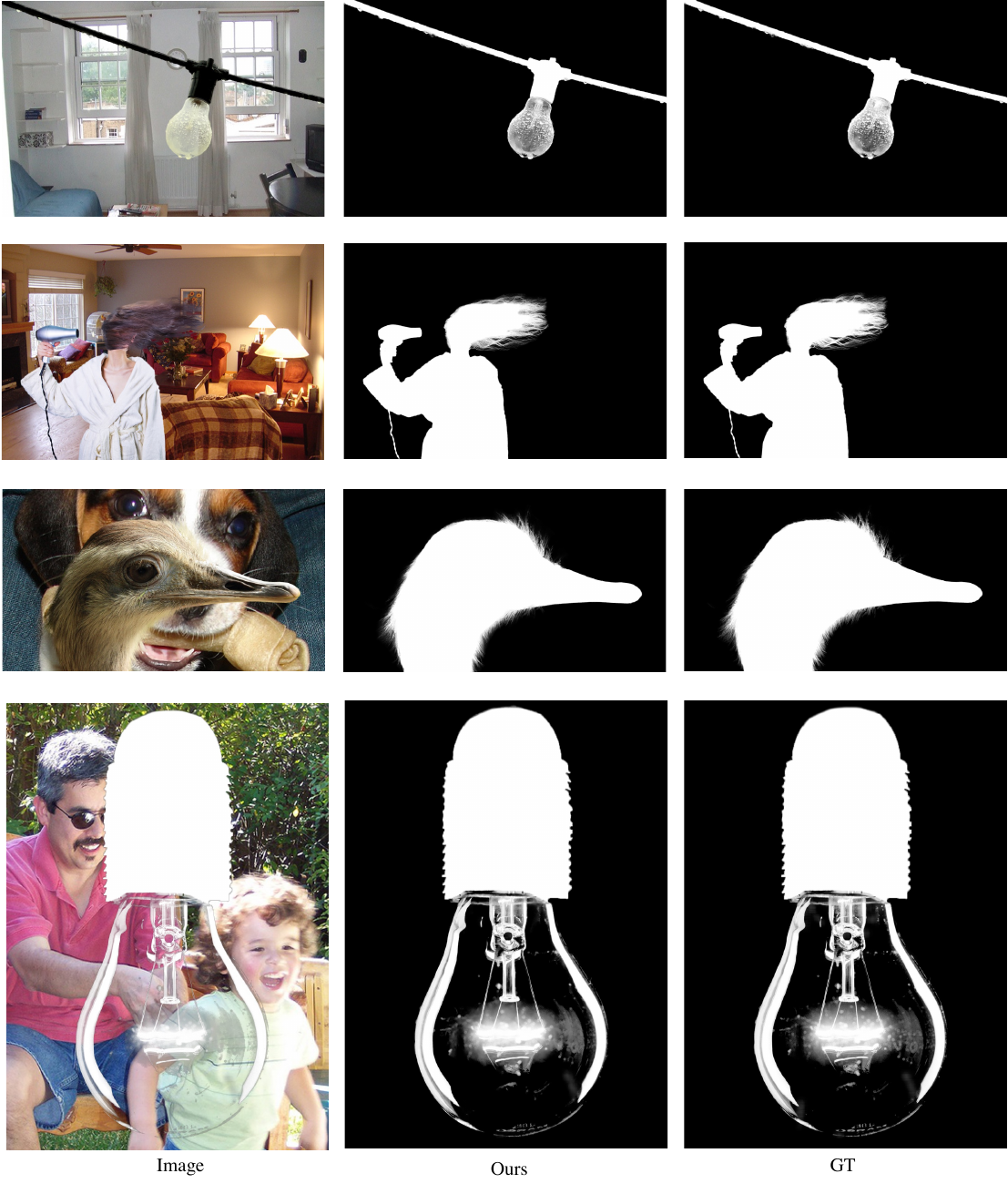}
  \caption{Qualitative results of our method on the Composition-1k test set.}
    \label{comp_supp2.examp}
 \end{figure*}

\begin{figure*}
  \centering
  \includegraphics[width=\textwidth]{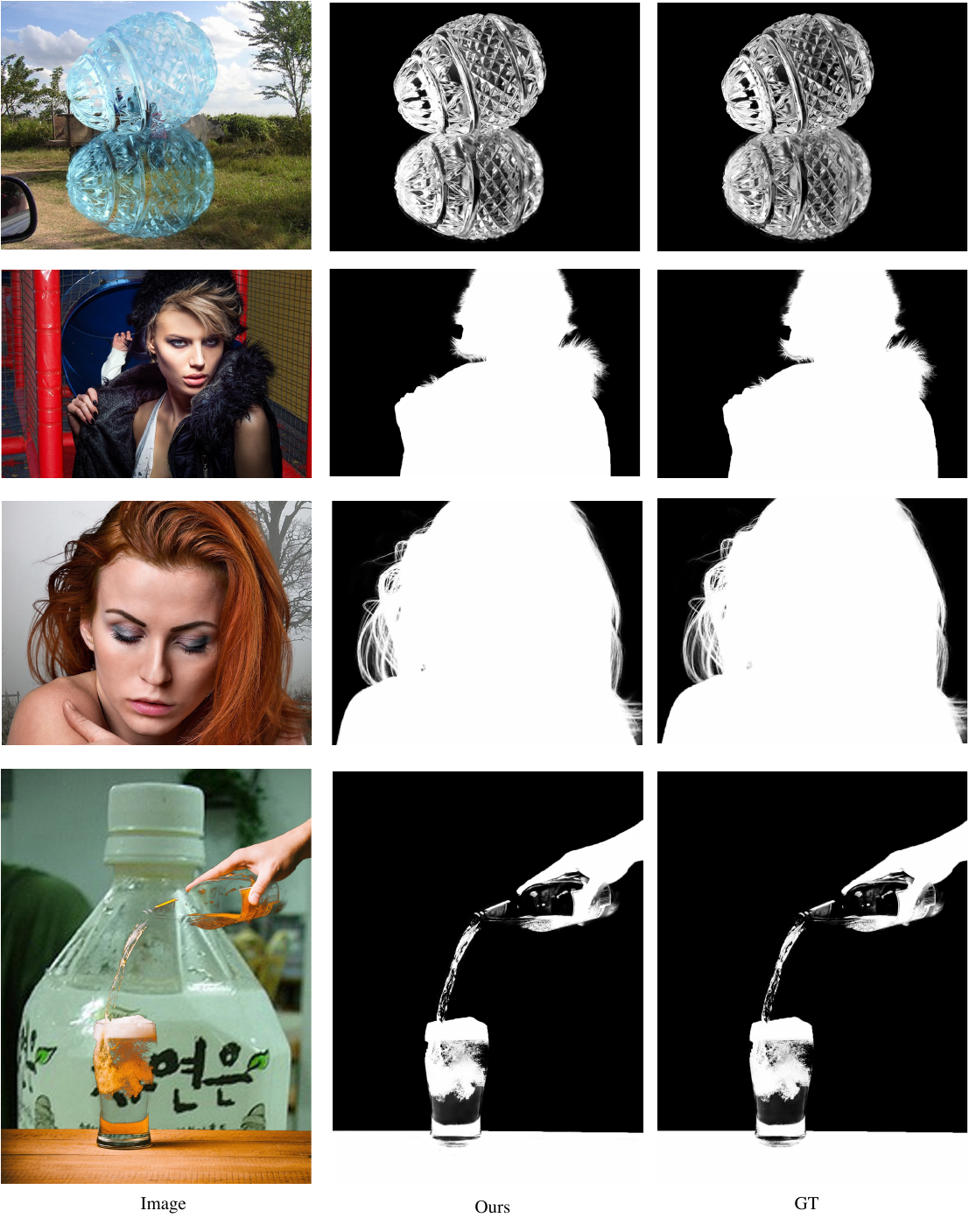}
  \caption{Qualitative results of our method on the Composition-1k test set.}
    \label{comp_supp3.examp}
 \end{figure*}

\begin{figure*}
  \centering
  \includegraphics[width=\textwidth]{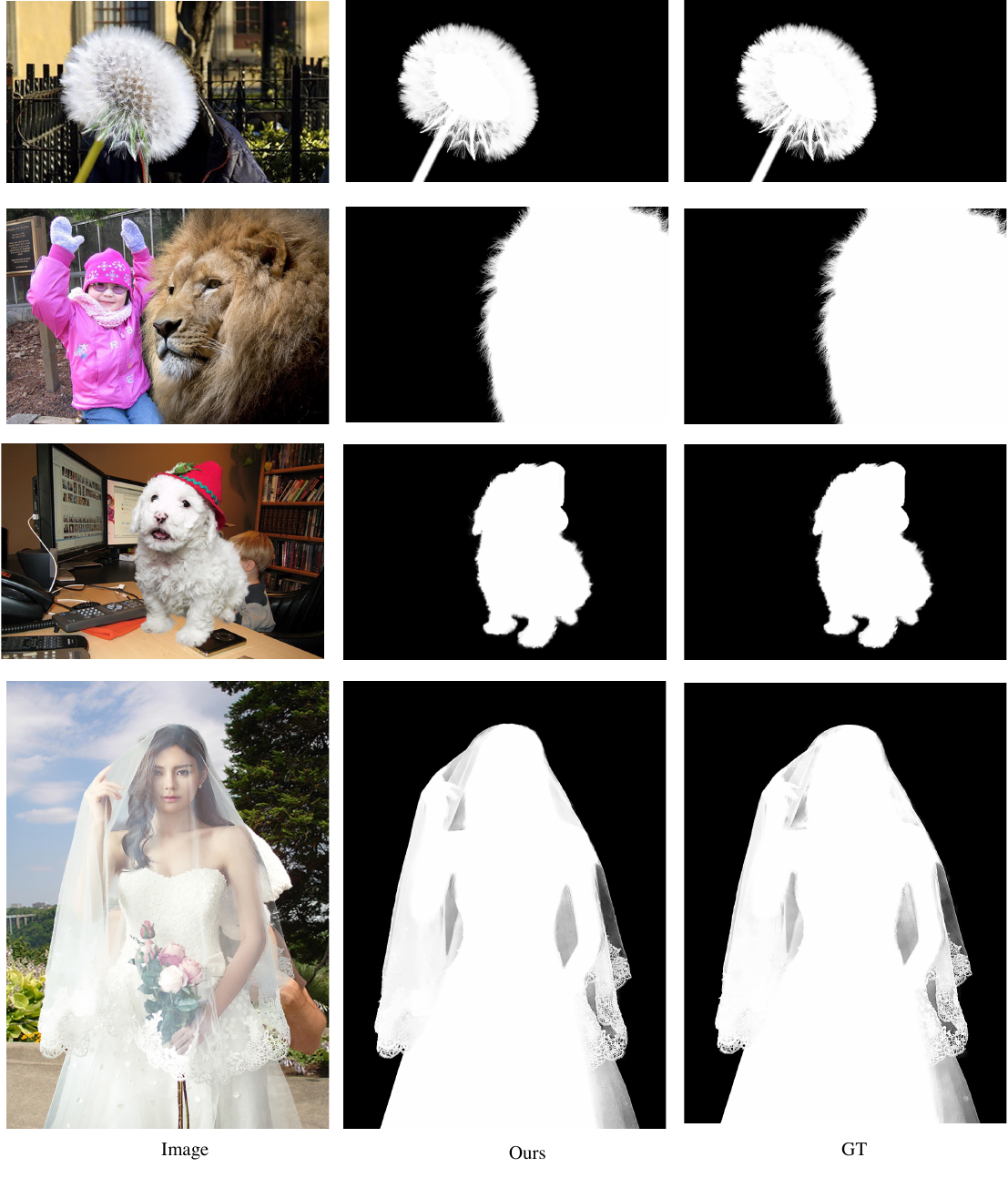}
  \caption{Qualitative results of our method on the Composition-1k test set.}
    \label{comp_supp4.examp}
 \end{figure*}

We fine-tune on Composition-1k with \MethodName~ initialization. As Fig.~\ref{comp_supp1.examp}, Fig.~\ref{comp_supp2.examp}, Fig.~\ref{comp_supp3.examp}, Fig.~\ref{comp_supp4.examp} show, we provide more qualitative results of our \MethodName~ on the Composition-1k test set. From the results, it can be seen that our method performs well in various detailed foreground objects. 

\vspace{-3mm}
\section{More fine-tuning qualitative results on Distinct-646}
\begin{figure*}
  \centering
  \includegraphics[width=\textwidth]{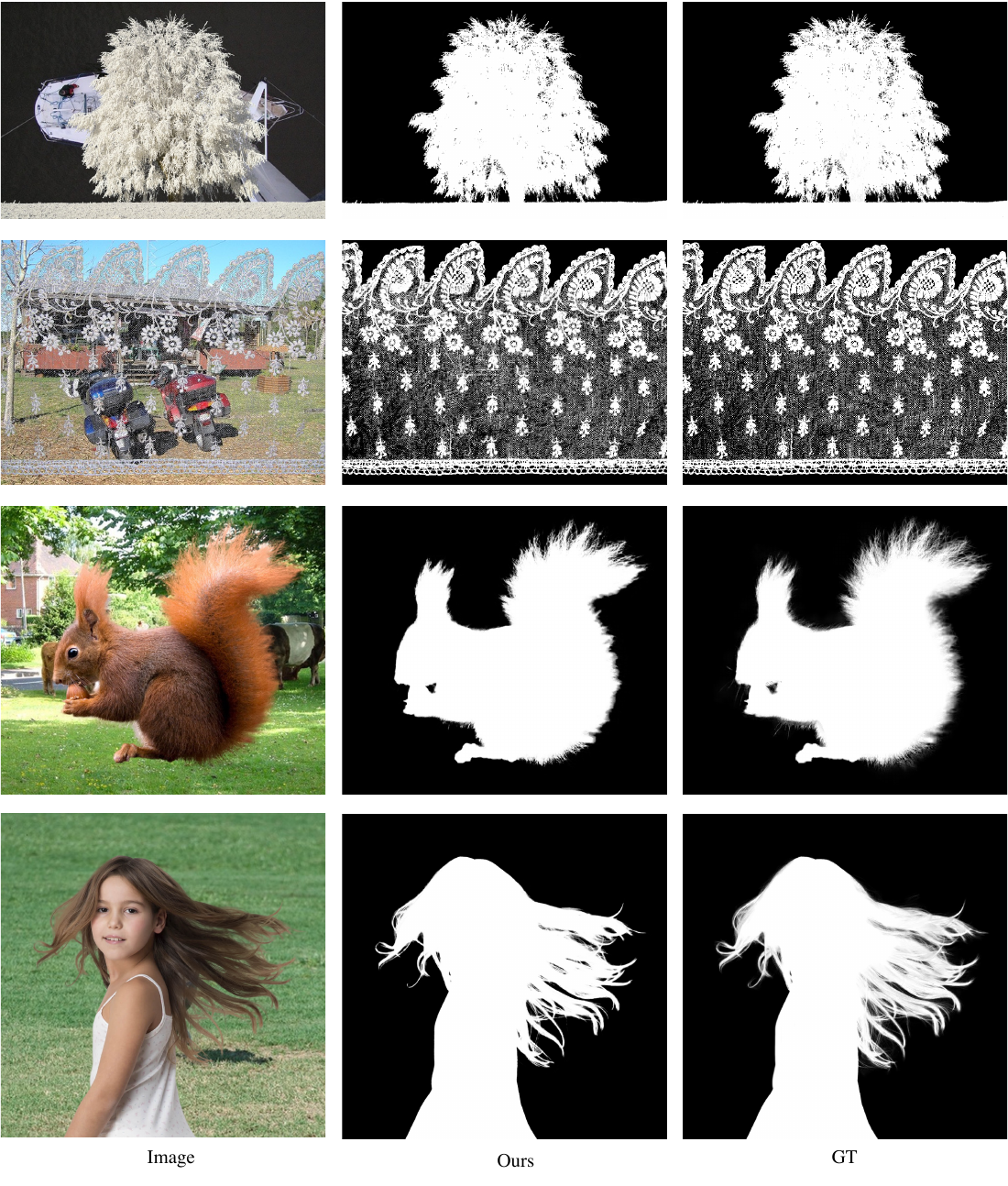}
  \caption{Qualitative results of our method on the Distinct-646 test set.}
    \label{dis_supp1.examp}
 \end{figure*}

\begin{figure*}
  \centering
  \includegraphics[width=0.9\textwidth]{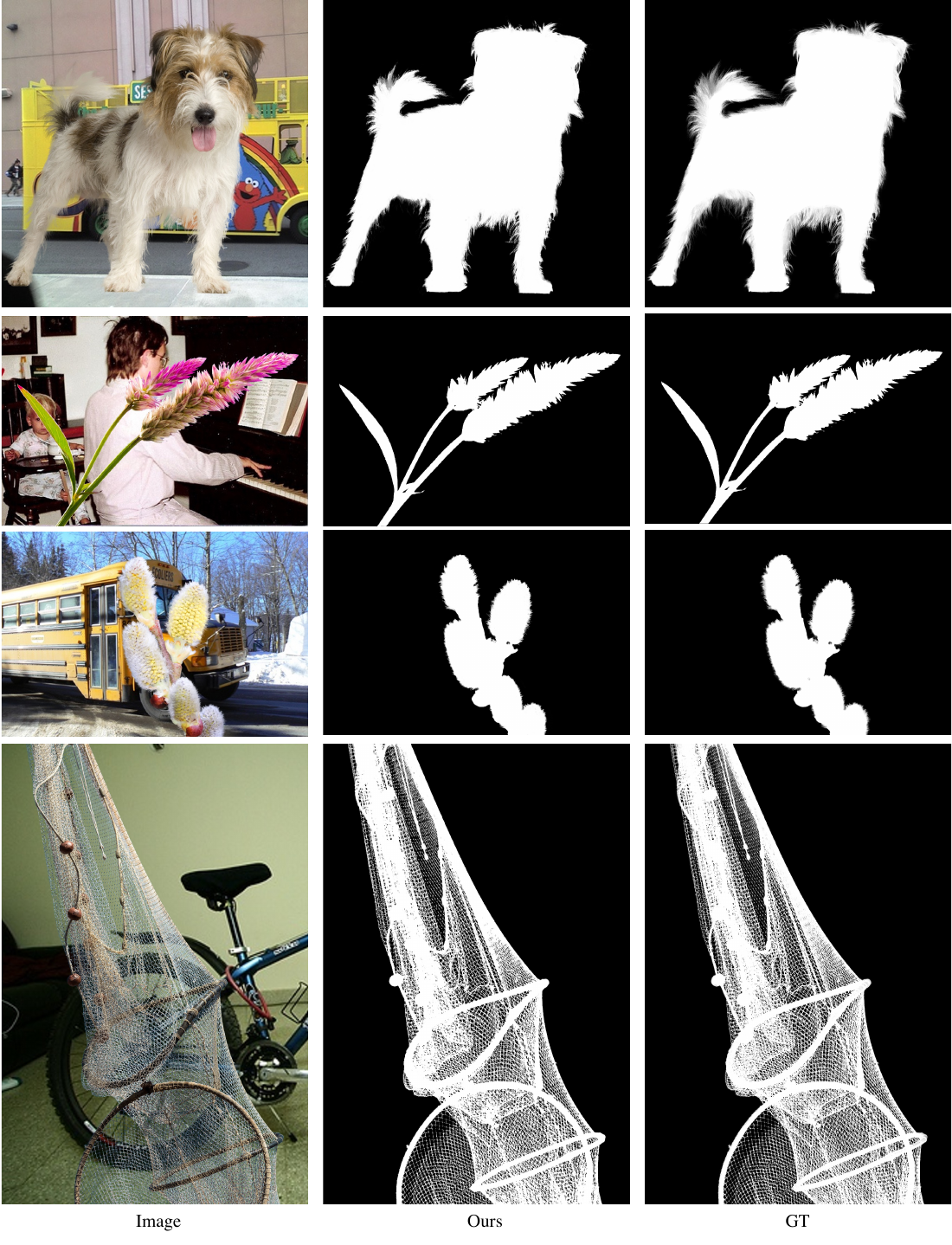}
  \caption{Qualitative results of our method on the Distinct-646 test set.}
    \label{dis_supp2.examp}
 \end{figure*}

\begin{figure*}
  \centering
  \includegraphics[width=\textwidth]{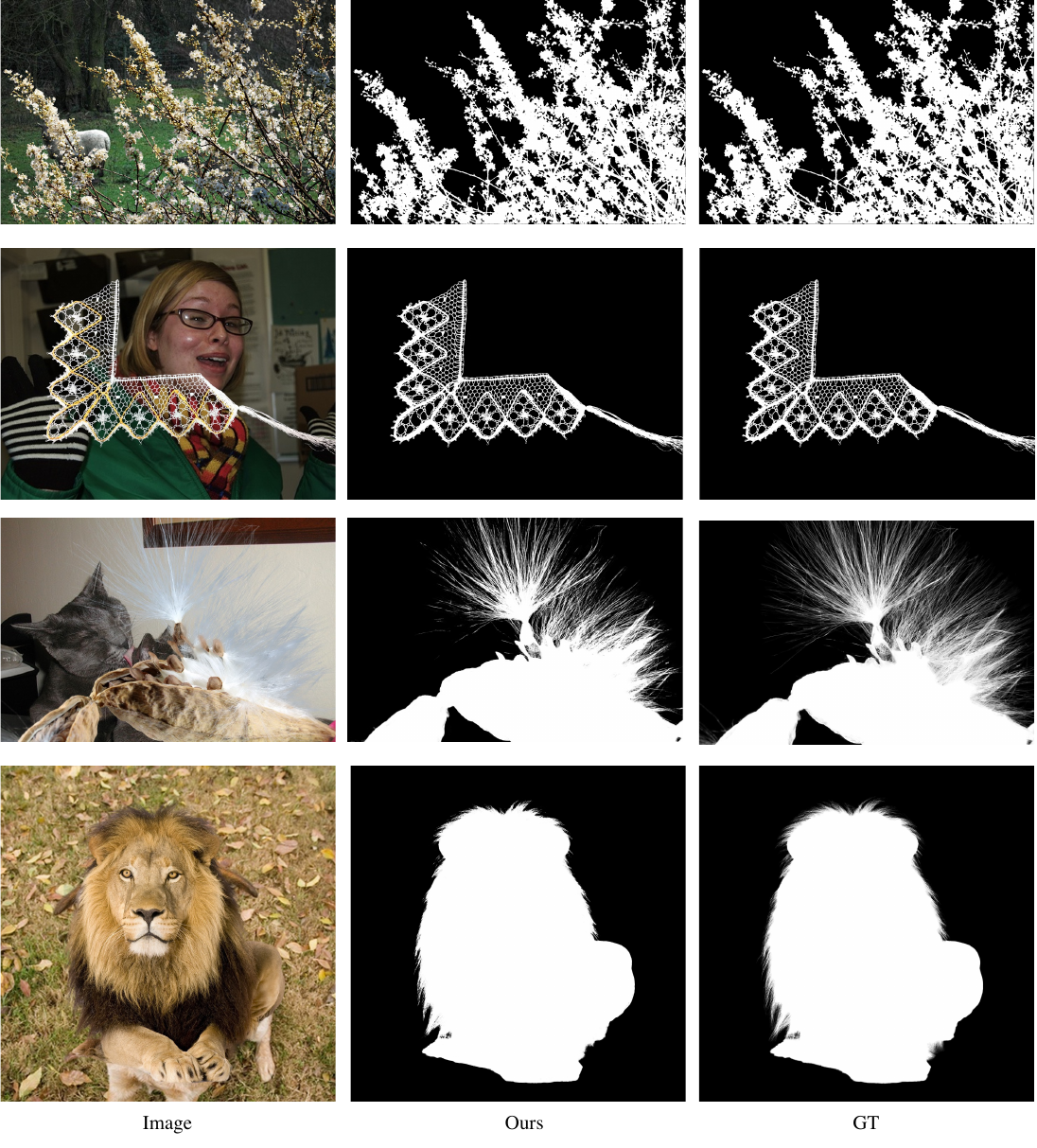}
  \caption{Qualitative results of our method on the Distinct-646 test set.}
    \label{dis_supp3.examp}
 \end{figure*}

\begin{figure*}
  \centering
  \includegraphics[width=0.9\textwidth]{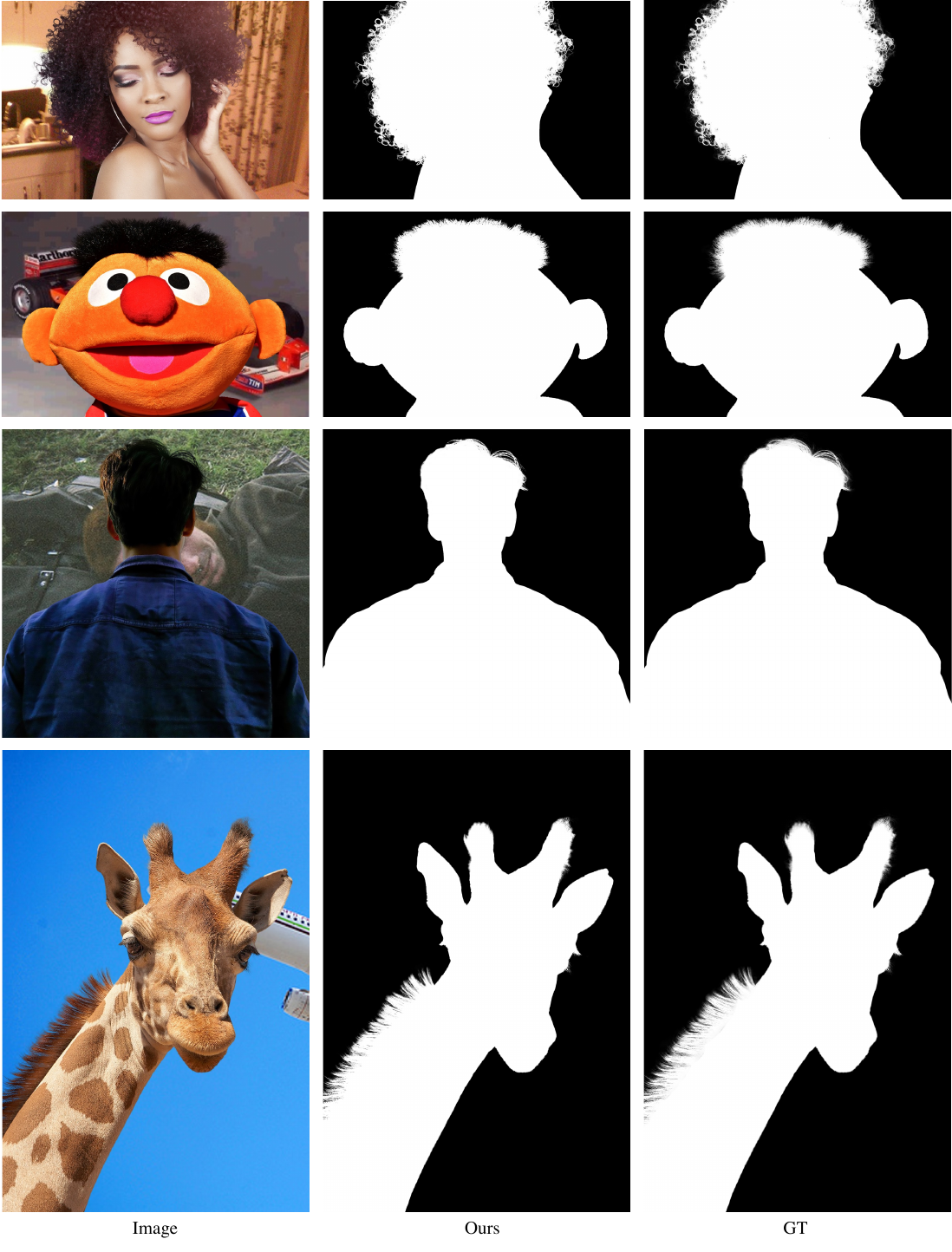}
  \caption{Qualitative results of our method on the Distinct-646 test set.}
    \label{dis_supp4.examp}
 \end{figure*}

 As Fig.~\ref{dis_supp1.examp} -- Fig.~\ref{dis_supp4.examp} show, we further present more qualitative results of our \MethodName~ on Distinct-646~\cite{qiao2020attention}. We first generate trimap by randomly dilating alpha mattes from the ground truth alpha matte with a threshold of 20 (following ~\cite{yu2021mask}). Then we use the fine-tuning model on Composition-1k to test directly on Distinct-646. We can see that our \MethodName~ achieves good performance, especially in detailed regions. 

\vspace{-3mm}
\section{More fine-tuning qualitative results on Natural human images}
\begin{figure*}
  \centering
  \includegraphics[width=0.9\textwidth]{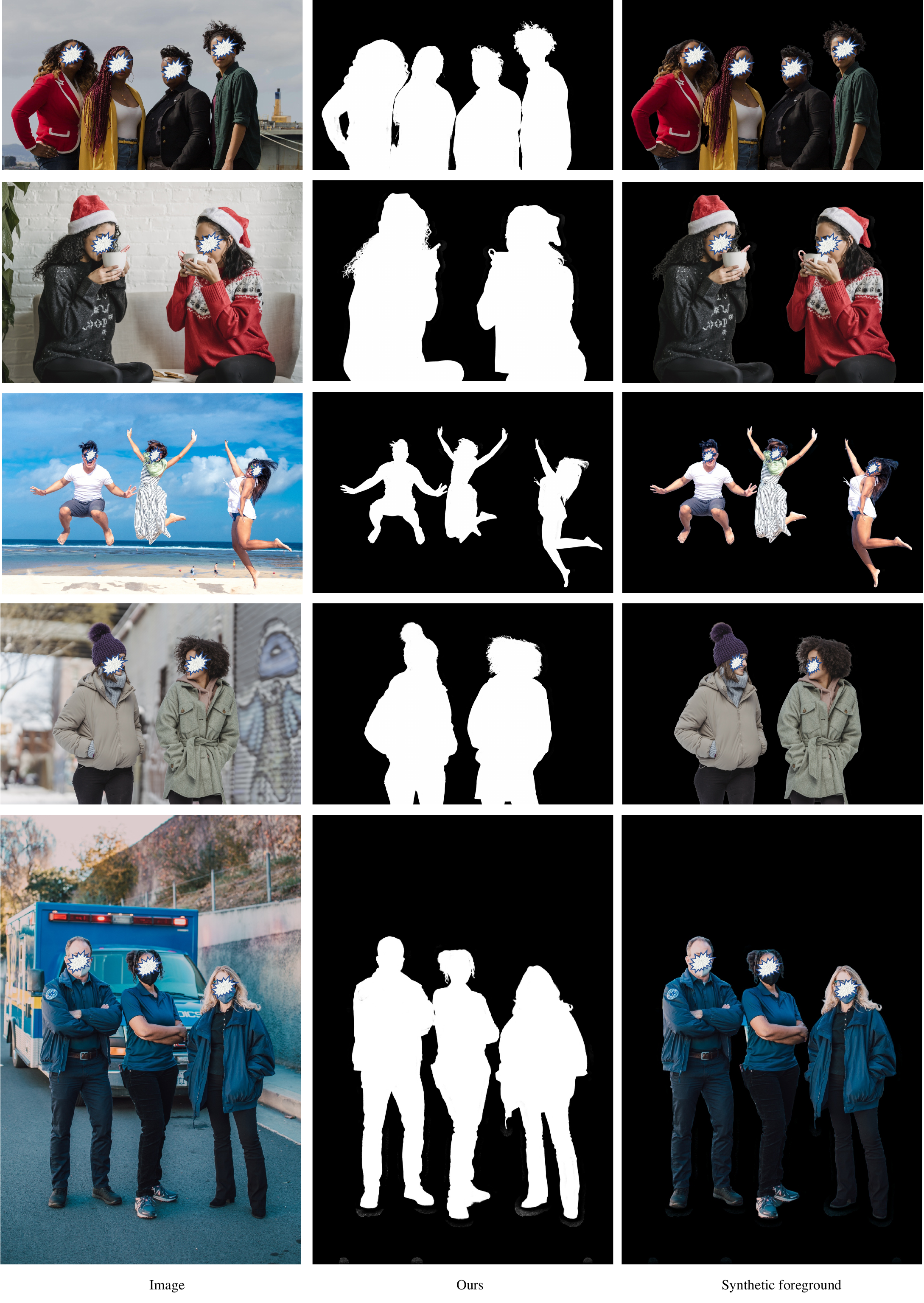}
  \caption{Qualitative results of our method on natural human images.}
    \label{him_supp1.examp}
 \end{figure*}

  To show the robustness of our method, we use a fine-tuning model on Composition-1k to test on 2k resolution natural human images as shown in Fig.~\ref{him_supp1.examp}. From left to right, we provide the image, generated alpha matte $\alpha$, and the synthetic foreground generated by $\alpha$ and image. We can see that our method performs well at extracting the boundary of the human body, such as the hair part.
 
